\documentclass[letterpaper, 10 pt, conference]{inc/ieeeconf}  

\IEEEoverridecommandlockouts                              

\usepackage{graphics} 
\usepackage{graphicx}
\usepackage{amsmath} 
\usepackage{amssymb}  
\usepackage{xcolor}
\usepackage{lipsum}
\newcommand{\todo}[1]{\textbf{\textcolor{red}{#1}}}

\usepackage{multirow}
\usepackage{hyperref}
\hypersetup{
    colorlinks = true,
    urlcolor = {cyan}
}
\usepackage{caption}
\usepackage{subcaption}
\usepackage{algpseudocode}
\usepackage{algorithm}
\usepackage{flushend}
\usepackage{booktabs}
\usepackage[T1]{fontenc}

\newif\ifevalnew
\evalnewtrue

\title{\LARGE \bf
Differentially Encoded Observation Spaces\\for Perceptive Reinforcement Learning
}

\author{Lev Grossman$^{1}$ and Brian Plancher$^{2}$
\thanks{$^{1}$Lev Grossman is with Berkshire Grey, Bedford, MA, USA. {\tt\footnotesize lev.grossman@berkshiregrey.com}}%
\thanks{$^{2}$Brian Plancher is with Barnard College, Columbia University, New York, NY, USA. {\tt\footnotesize bplancher@barnard.edu}}%
}

\begin{document}
\maketitle
\thispagestyle{empty}
\pagestyle{empty}


\begin{abstract}
Perceptive deep reinforcement learning (DRL) has lead to many recent breakthroughs for complex AI systems leveraging image-based input data. Applications of these results range from super-human level video game agents to dexterous, physically intelligent robots. However, training these perceptive DRL-enabled systems remains incredibly compute and memory intensive, often requiring huge training datasets and large experience replay buffers. This poses a challenge for the next generation of field robots that will need to be able to learn on the edge in order to adapt to their environments. In this paper, we begin to address this issue through differentially encoded observation spaces. By reinterpreting stored image-based observations as a video, we leverage lossless differential video encoding schemes to compress the replay buffer without impacting training performance. We evaluate our approach with three state-of-the-art DRL algorithms and find that differential image encoding reduces the memory footprint by as much as 14.2$\times$ and 16.7$\times$ across tasks from the Atari 2600 benchmark and the DeepMind Control Suite (DMC) respectively. These savings also enable large-scale perceptive DRL that previously required paging between flash and RAM to be run entirely in RAM, improving the latency of DMC tasks by as much as 32\%.


\end{abstract}

\section{INTRODUCTION}
With its ability to synthesize complex behaviors in both simulated and real environments, deep reinforcement learning (DRL) has been applied to solve a host of robotic-specific problems ranging from dexterous manipulation~\cite{Andrychowicz18}, to quadrupedal locomotion~\cite{Miki22}, to high-speed drone racing~\cite{Song21}, as well as more general artificial intelligence (AI) feats of mastering tabletop~\cite{silver2017mastering} and competitive video games such as Dota 2~\cite{openai2019dota} and Minecraft~\cite{hafner2023mastering}.

Despite these achievements, DRL remains very sample inefficient, often requiring massive amounts of training data to learn. Because much of this persistent data is loaded into experience replay buffers during training, DRL is extremely memory intensive, limiting the number of computational platforms that can support such operations, and largely confining state-of-the-art model training to the cloud.

Interestingly, it is the observations in the replay buffer which often consume the majority of the memory, reaching over 90\% for proprioceptive models~\cite{grossman23JustRound}. The recent interest in perceptive DRL models~\cite{hafner2023mastering, yarats2021DrQ-v2} only further exacerbates this problem. For example, as compared to the 224 bytes needed to store the proprioceptive observation space for the DeepMind Control Suite (DMC)~\cite{tassa2018deepmind} quadruped, a single 84x84 grayscale image requires over 7kB of memory.
Consequently, deployment to the edge is unrealistic for such approaches. Still, many robots, especially those involved in tasks as consequential as search-and-rescue and space exploration~\cite{Gankidi17}, will have to adapt to ever-changing environmental conditions and continue to optimize and update their internal policies over the course of their lifetime~\cite{Thrun95}, often in remote areas without access to fast network connections. As such, approaches that reduce the overall memory footprint of perceptive DRL are needed to enable edge deployments.
 
In this paper, we begin to address this issue through differentially encoded observation spaces. That is, by reinterpreting stored image-based observations as a video, we leverage lossless differential video encoding schemes to compress the replay buffer without impacting training performance. 
%
We evaluate our approach across ten Atari 2600 benchmark tasks~\cite{Bellemare_2013ALE} as well as two robotic control tasks from within the DMC, using three state-of-the-art DRL algorithms. We find that differential image encoding reduces the memory footprint by as much as 14.2$\times$ and 16.7$\times$ for the Atari and DMC tasks respectively. These savings also enable large-scale DRL that previously required paging between flash and RAM to be run entirely in RAM, improving latency for the DMC tasks by as much as 32\%. We release our software and experiments open-source at: \href{https://github.com/A2R-Lab/DiffCompressDRL}{\texttt{github.com/A2R-Lab/DiffCompressDRL}}.

\section{RELATED WORK}
\textbf{Perceptive DRL:} Perceptive or image-based control and deep reinforcement learning (DRL) share common roots. First proposed a decade ago, Deep Q-Networks (DQNs)~\cite{mnih2013playing} bridged the gap between deep learning and RL, achieving super-human performance on many Atari 2600 video games using only visual input. Many model-free, DQN-based variants have followed~\cite{hessel2017rainbow, dabney2017distributional}, further improving game-playing performance directly from pixel observations. Still, these early model-free methods remained quite data expensive. To address this, some turned to model-based or world-model learning in both the Atari~\cite{kaiser2020modelbasedSimPLe} as well as robotic control~\cite{hafner2023mastering,hafner2019learningPlaNet} domains. More recently, model-free methods have improved sample efficiency further through contrastive learning~\cite{srinivas2020curl} and image augmentation~\cite{kostrikov2021DrQ}. Despite these improvements, perceptive DRL remains memory intensive, owing especially to the large size of image observations. For memory constrained systems and more complex tasks that require larger replay buffers, it has become necessary to store observations directly on disk, increasing the average memory access time and thus overall model training times~\cite{yarats2021DrQ-v2}.



\textbf{Compressed Perceptive DRL:} While recent image-based, model-free DRL methods have improved performance and sample efficiency through data augmentation~\cite{laskin2020reinforcementaugment}, contrastive learning~\cite{srinivas2020curl}, and more robust latent observation space representations~\cite{yarats2021reinforcementrepresentations}, little work has been done (aside from moving from RGB to grayscale~\cite{laskin2020reinforcementgrayscale}) to reduce the size of the stored image observations. Model-based methods offer an alternative by learning a world-model directly, thereby reducing the dependence on large amounts of replay experience~\cite{Hafner18, hafner2022mastering}. While these methods do show promise in highly complex simulated environments~\cite{hafner2023mastering} as well as in some real laboratory settings~\cite{wu2022daydreamer}, world-model learning tends to still struggle when the underlying world dynamics become highly complex~\cite{tomar2021learning}. For this reason, model-free methods remain popular for robotics applications leveraging learning in the real world~\cite{pmlr-v164-yu22a}.





\textbf{Video Encoding for Learning:} 
Traditionally, video encoding schemes aim to reduce the visual redundancy in digital video files without impacting human-perceived quality~\cite{wood2022task}. Modern video encoding standards such as MPEG-4~\cite{MPEG-4} and H.264~\cite{chen2006introductionH264} use both intra- and inter-picture compression techniques to achieve this result. With the recent proliferation of deep learning for computer vision, some have begun optimizing traditional video encoding standards for learning frameworks~\cite{yang2021videoVCM}. While these methods have been applied to classification~\cite{huang2021visualVCSclassification}, object detection~\cite{Yang_2020VCSdetection}, and instance segmentation~\cite{Le_2021VCSInstance} tasks among others, little has been done to incorporate video compression within DRL.





\section{BACKGROUND} \label{sec:background}
\subsection{Reinforcement Learning}
Reinforcement learning poses problems as Markov decision processes (MDPs), where an MDP is defined by a set of observed and hidden states, $\mathcal{S}$, actions, $\mathcal{A}$, stochastic dynamics, $p(s_{t+1}|s_t, a_t)$, a reward function $r(s,a)$, and a discount factor, $\gamma$. The RL objective is to compute the policy, $\pi^*(s,a)$, that maximizes the expected discounted sum of future rewards, $\mathbb{E}_{s,a}\left(\sum_t \gamma^t r_t \right)$.

Perceptive RL, where states $s$ are represented by images, adds an additional caveat to the RL problem formulation, as representing a state using a single image rendering of the system is often not sufficiently descriptive~\cite{yarats2021DrQ-v2}. To alleviate this issue, states are approximated by concatenating together the $f$ immediately preceding image frames $s_t = \{ o_{t - f + 1}, \dots, o_{t}\}$, where $o_t$ is the image observation at timestep $t$~\cite{mnih2013playing}.

DRL algorithms parameterize the solution to these large MDPs with neural networks. In this work, we leverage three different state-of-the-art DRL algorithms: Proximal Policy Optimization~\cite{Schulman17}, Quantile Regression DQN~\cite{dabney2017distributional}, and Data-regularized Q-v2~\cite{yarats2021DrQ-v2}. In the remainder of this section, we provide more detail on each algorithm.

\subsubsection{Proximal Policy Optimization (PPO)}

Proximal Policy Optimization~\cite{Schulman17} is an on-policy, policy gradient~\cite{Sutton00} algorithm that employs an actor-critic framework to learn both the optimal policy, $\pi^*(s,a)$, as well as the optimal value function, $V^*(s)$. Both the policy $\pi_\theta$ and value function $V_\phi$ are parameterized by neural networks with weights $\theta$ and $\phi$ respectively. During training, PPO needs to store the parameters of both its value and policy networks\footnote{Many PPO implementations save space by using one central MLP with two additional single-layer policy and value function model heads. We focus on the standard PPO model approach in this section for clarity.}--usually shallow multilayer perceptrons (MLPs)--as well as its on-policy rollout buffer $\mathcal{D}_k$. $\mathcal{D}_k$ stores $(s, a, r)$ tuples, which are refreshed during each iteration of the algorithm.

\subsubsection{Quantile Regression DQN (QR-DQN)}

Quantile Regression DQN~\cite{dabney2017distributional} improves upon the highly influential Deep Q-Networks (DQN)~\cite{mnih2013playing} algorithm, which uses experience replay to learn an optimal Q-function approximated by a neural network $Q_\theta$. While in the original DQN paper the authors train using a Huber loss~\cite{huber64} to maximize mean return, QR-DQN uses a custom quantile Huber loss in order to learn a distribution over expected returns. With this change, QR-DQN is able to significantly outperform DQN on a benchmark of Atari 2600 games. Unlike PPO, QR-DQN stores collected experience across all iterations in a much larger replay buffer $\mathcal{D}$, functioning as a size limited queue.

\subsubsection{Data-regularized Q-v2 (DrQ-v2)}

Data-regularized Q-v2~\cite{yarats2021DrQ-v2} improves upon the Data-regularized Q~\cite{kostrikov2021DrQ} model-free RL algorithm. Both algorithms use data augmentation in conjunction with an actor-critic learning method in order to perform image-based continuous control. DrQ-v2 applies image augmentation through random shifts and then passes the augmented images through an encoder network $f_\xi$ to produce the final observations, which are used by an actor-critic DDPG~\cite{Lillicrap15} setup to learn the solution policy. DrQ-v2 achieves state-of-the-art model-free results on the DeepMind Control Suite, rivaling the performance of the popular model-based DreamerV2~\cite{hafner2022mastering}, while doing so 4$\times$ faster (in terms of wall-clock training time). Like QR-DQN, DrQ-v2 stores its collected experience data in a large replay buffer $\mathcal{D}$.


\begin{figure*}[!t]
     \centering
     \vspace{5pt}
     \includegraphics[width=0.7\textwidth]{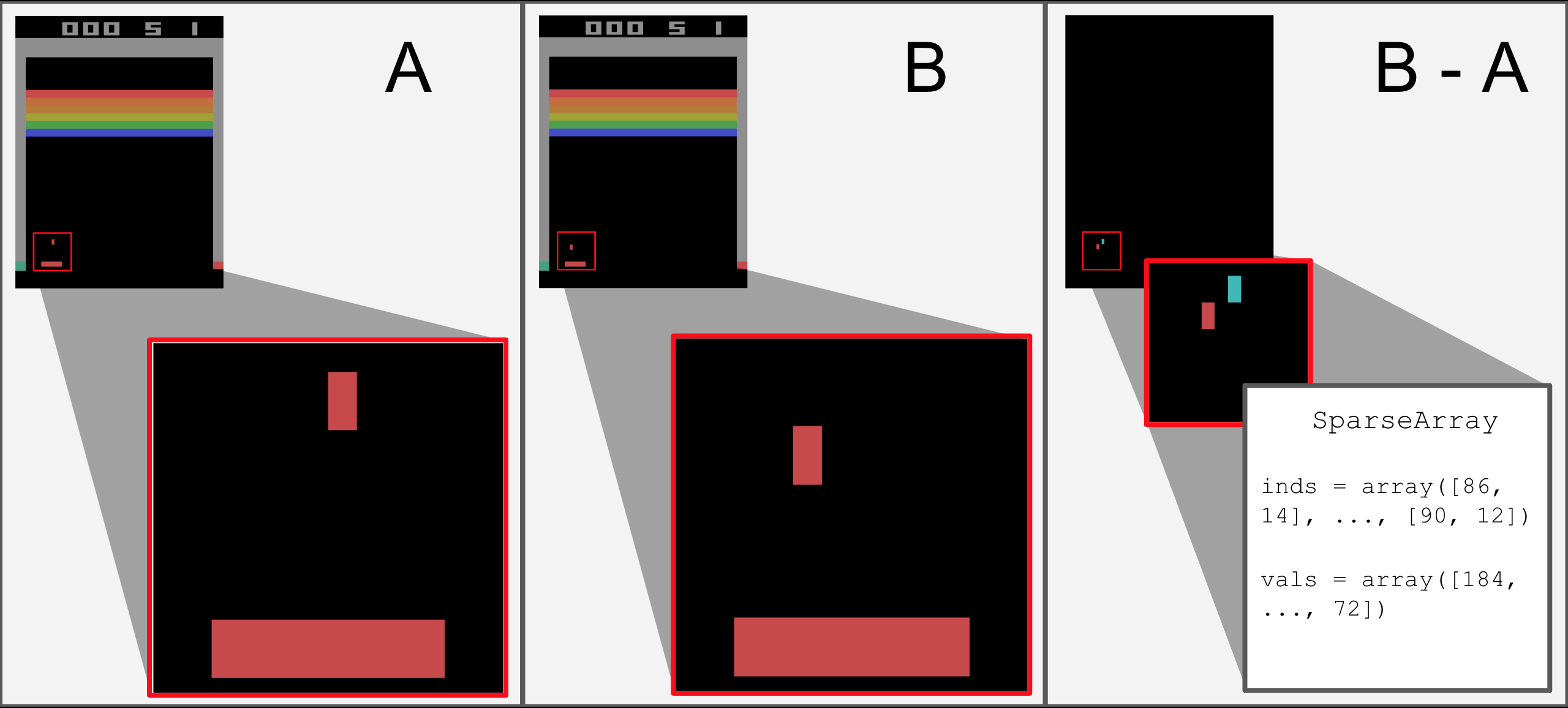}
     \caption{A graphical example of a differential image encoding taken from the Breakout Atari 2600 environment~\cite{Bellemare_2013ALE}. The two images, A and B shown on the left and middle, are very similar. The only difference in the resulting B-A image, shown on the right, occurs for the two different locations of the falling object. We can therefore encode this small difference in a highly compressed sparse matrix format.}
     \label{fig:image_diff}
     \vspace{-15pt}
\end{figure*}

\subsection{Memory Requirements for Perceptive DRL}

Previous work showed that for proprioceptive DRL, the memory requirements for PPO were dominated by the observation space, consuming over 90\% of the total memory~\cite{grossman23JustRound}. Perceptive DRL does not change this and actually only makes matters worse. For example, the DMC quadruped requires only 224 bytes for a single proprioceptive observation, while a single 84x84 grayscale image requires over 7kB of memory. For the off-policy QR-DQN as well as DrQ-v2, an even greater percentage of the overall memory requirement is dominated by the observation space since, as mentioned above, their replay buffers are generally orders of magnitude larger than PPO's rollout buffer.

\subsection{Differential Video Encoding}

Modern video coding formats and codecs, like the H.264~\cite{chen2006introductionH264} standard, make use of a number of compression and quantization techniques in order to reduce the overall size of video streams while preserving quality. One of the compression techniques employed is differential encoding (DE)~\cite{IntroDataCompCh11DE}, which can be used to compress both single images as well as multiple subsequent image frames by encoding the difference in inter-pixel or pixel-block values/intensities. In this work, we make use of a simplified version of differential video encoding--with inspiration from standard text-based DE used by most version control systems~\cite{ruparelia2010historyVCS}--in order to compress the pixel-based observations found in rollout and replay buffers. We present our encoding scheme in detail in the following section.

\section{METHOD}
\label{sec:method}

In this section, we detail a custom, lossless compression technique for image-based observation spaces. While we later show empirical results using PPO, QR-DQN, and DrQ-v2, we note that this differential encoding-based scheme can be applied to any image-based RL method in order to reduce the size of its stored observations.

\subsection{Differential Encoder}
During DRL training, there tends to be a level of temporal sortedness in experience replay buffers~\cite{Lin92ExperienceReplay, schaul2016prioritized}, owing to the fact that many model-free RL algorithms add experience $(s, a, r)$ sequentially as they train. Therefore, while two states $s_i, s_j$ drawn at random from a replay buffer may differ dramatically, two adjacent states $s_{i-1}, s_i$ have a high likelihood of being similar, as they may originate from the same episode, just one control step apart. Applied to image-based observations, this means that images $o_{i-1}$ and $o_i$ will tend to be visually quite similar.

This similarity between adjacent image frames allows for a natural differential encoding-based compression scheme (Figure~\ref{fig:image_diff}), which takes the difference between $o_i - o_{i-1}$ and stores the result in sparse matrix format $S_i = \texttt{SparseArray}(o_i - o_{i-1})$. To decode $o_i$, it is then only necessary to store $o_{i-1}$ and $S_i$, as:
\begin{equation}
    o_i = o_{i-1} + S_i.
    \label{eq:decode_sparse}
\end{equation}

In order to efficiently store and access these image differences, we define a custom sparse matrix implementation $\texttt{SparseArray}$ that stores all nonzero pixels using two arrays $\texttt{inds}$ and $\texttt{vals}$, where $\texttt{inds}$ stores pixel locations and $\texttt{vals}$ stores pixel values. Because image observations in both the Atari 2600 benchmark and DeepMind Control Suite are stored as 84x84, 8-bit unsigned integer arrays (assuming grayscale images), we let $\texttt{inds}$ store 8-bit unsigned integers and $\texttt{vals}$ store 16-bit signed integers. Assuming $o_i - o_{i-1}$ results in $n$ nonzero pixel values, we can store a 7kB input image in $4n$ bytes. 

While it is possible that $o_i - o_{i-1}$ results in an image with a majority nonzero pixel values, in practice we find the number of nonzero pixels $n$ to be small. Nevertheless, we handle this corner case explicitly in $\texttt{SparseArray}$. If $4n > 84\times84$, we store the full image array in order to cap the maximum image size at 7kB.

\subsection{Observation Indexing} \label{sec:obs_ind}

As noted in the previous section, image-based learning methods generally store states $s$ as a stack of the $f$ prior image observations. Letting $f=4$ (as is the case for our Atari 2600 experiments in Section~\ref{sec:results}), this results in each state taking up $84\times84\times4 = 28$kB of memory.

Instead of storing states in this full image stack format, we define two separate arrays $\texttt{obs}$, which stores the individual image observations, and $\texttt{obs\_inds}$, which stores pointers into $\texttt{obs}$ for each state. To reconstruct the full state $s_i$, we simply concatenate the individual observations referenced by $\texttt{obs\_inds}_i$. In most cases $\texttt{obs\_inds}_i = \{(i-f+1), (i-f+2), \dots , i \}$ resulting in:
\begin{equation}
    s_i = \{\texttt{obs}_{(i-f+1)}, \texttt{obs}_{(i-f+2)}, \dots, \texttt{obs}_{i}\}.
    \label{eq:obs_inds}
\end{equation}
However, in various corner cases, this is not the case, necessitating the use of the pointer array. For example, for the first few observations, at the start of an episode, individual observations need to be repeated to fill a history of size $f$.

Through observation indexing alone, we can reduce the memory required to store all states by almost $f\times$, as $f$ pointers are a fraction of the size of the $(f-1)$ image observations we no longer need to save. 

\subsection{Observation Compressor} \label{sec:obs_comp}

With both the differential image encoder and smart observation indexing, we can define an observation compressor to handle the storage and retrieval of image-based observations. Following best practices from the video encoding literature, we apply our approach to each set of $f$ images.

Our compressor, $\texttt{ObsComp}$, internally stores three arrays: $\texttt{obs}$, $\texttt{sparse\_obs}$, and $\texttt{obs\_inds}$.  Using these arrays we can then define the following two operations: $\texttt{ObsComp.get(i)}$ and $\texttt{ObsComp.set(s, i)}$, which get and set the uncompressed state $s_i$ respectively.

To set a state $s_i$, we first compute the observation index array $\texttt{obs\_inds}_i = \{(i - f + 1), \dots, i\}$. Next, we check if $i \mod f = 0$, and if so store the raw image frame $\texttt{obs}_{i / f} = o_i$. If not, we compute and store the $\texttt{SparseArray}$, $\texttt{sparse\_obs}_{i / f, (i \mod f) - 1} = o_i - \texttt{obs}_{i - i \mod f}$.

To get a state $s_i$, we retrieve the observation indices $\texttt{obs\_inds}_i$ and concatenate the individual frames referenced either directly from $\texttt{obs}$ or from decoding compressed frames stored in $\texttt{sparse\_obs}$ (Equation~\ref{eq:decode_sparse}).

\subsection{Theoretical Compression Factor}

With these definitions in mind, we can evaluate the theoretical compression factor for our approach, that is the reduction ratio in overall memory consumption. 

Let $\texttt{obs}$ store uncompressed image observations $o_i$ and be of shape $(d, I, I)$, where $d = |\mathcal{D}| / f$ is the size of the uncompressed replay buffer divided by the frame stack size $f$, and $(I, I)$ is the size of a single grayscale input image.

Let $\texttt{sparse\_obs}$ store compressed image observations $S_i$ in $\texttt{SparseArray}$ format and be of shape $(d, f - 1)$. Note that each element of this array contains the overhead to store a pointer to a $\texttt{SparseArray}$ object, which can then be accessed to perform observation decompression (Equation~\ref{eq:decode_sparse}).


Finally, let $\texttt{obs\_inds}$ store frame stack indices (as explained in the previous sub-section) and be of shape $(|\mathcal{D}|, f)$. 

Assuming each $S_i$ incurs an overhead of an 8-byte pointer and $N$ is the number of total nonzero pixel values for all $S_i \in \texttt{sparse\_obs}$, we can define the memory size of $\texttt{ObsComp}$ as the following (in bytes):
\begin{equation}
    I^2d + 8d(f-1) + 4N + 4|\mathcal{D}|f.
    \label{eq:obs_comp}
\end{equation}
In comparison, to store all observations without any form of compression would take $|\mathcal{D}| \times I \times I \times f$ bytes. Overall, our method yields a theoretical compression factor of:
\begin{equation}
    \frac{I^2|\mathcal{D}|f}{I^2d + 8d(f - 1) + 4N + 4|\mathcal{D}|f}.
    \label{eq:comp_factor}
\end{equation}

Letting $N = d(f - 1)n$, where $n$ is the average number of pixels that need to be stored per compressed image, and $I=84$ (as used in our experiments in Section~\ref{sec:results}), we can further simplify the compression factor in terms of $f$ and $n$:
\begin{equation}
    \frac{1764f}{(1762-n)/f + 2 + n}.
\end{equation}

\begin{figure}[!h]
     \centering
     \includegraphics[clip, trim=1cm 19cm 7.5cm 1.5cm, width=0.9\columnwidth]{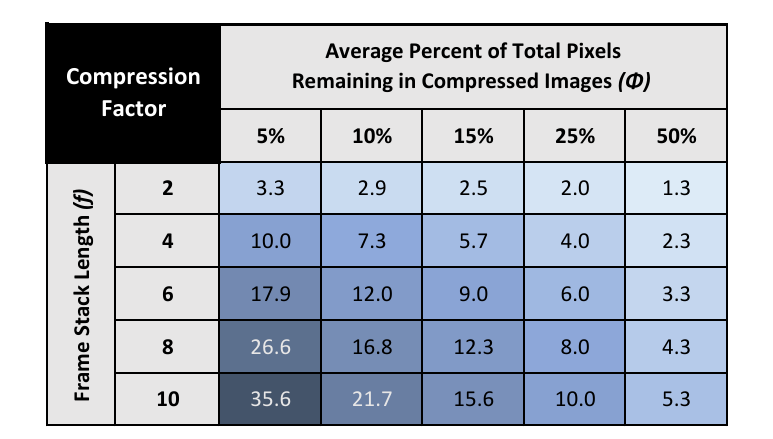}
     \caption{Theoretical compression factor for different values of frame stack length $f$ and average percent of total pixels remaining per compressed image $\phi$. Higher values correspond to more compression achieved.}
     \label{fig:compression}
     \vspace{-10pt}
\end{figure}

Noting that $n = I^2\phi = 7056\phi$, where $\phi$ is the average \emph{percentage} of pixels that need to be stored per compressed image, we plot the theoretical compression factor resulting from our approach for varying values of $f$ and $\phi$ in Figure~\ref{fig:compression}. We observe that even with a relatively small frame stack length of $f=4$, an order of magnitude compression factor can still be achieved as long as at most an average of 5\% of the pixels remain in the encoded images. Similarly, with a larger frame stack length of $f=10$, we can also still achieve an order of magnitude compression factor, even with up to 25\% of the pixels remaining. 

\section{EXPERIMENTS AND RESULTS} \label{sec:results}
\ifevalnew
    We evaluate the effectiveness of our differential encoder-based compression framework using tasks from both the Atari 2600 benchmark~\cite{Bellemare_2013ALE} as well as the DeepMind Control Suite (DMC)~\cite{tassa2018deepmind}. We analyze our results in terms of training speed, memory consumption (and associated compression factor), and convergence performance. Our full implementation, including values for all hyperparameters used in our experiments, can be found in our open-source GitHub repository at: \href{https://github.com/A2R-Lab/DiffCompressDRL}{\texttt{github.com/A2R-Lab/DiffCompressDRL}}.

\subsection{Methodology}

For all Atari benchmark tasks, we set $f$ to the default value of $4$ and train using two popular on- and off-policy reinforcement learning algorithms: Proximal Policy Optimization (PPO)~\cite{Schulman17} and Quantile Regression DQN (QR-DQN)~\cite{dabney2017distributional}. We use publicly available implementations of both PPO and QR-DQN (credit: \texttt{stable-baselines3}~\cite{stable-baselines3}), modifying only the buffer logic in order to accommodate observation compression. For the DeepMind Control Suite tasks, we set $f$ to the default value of $3$ and train on a state-of-the-art off-policy, data augmentation-based reinforcement learning algorithm: Data-regularized Q-v2 (DrQ-v2)~\cite{yarats2021DrQ-v2}. We use Meta Research's official implementation of DrQ-v2, again only modifying logic pertaining to the storage and retrieval of image observations. Unless otherwise noted, we run all experiments using the default PPO, QR-DQN, and DrQ-v2 hyperparameters. Exact values for all hyperparameters can be found on our GitHub repository.\footnote{Training was done using a high-performance workstation with a $3.2$GHz 16-core Intel i9-12900K and a $2.2$GHz NVIDIA GeForce RTX 4090 GPU running Ubuntu 22.04 and CUDA 12.1.}

\begin{table}[!t]
\vspace{10pt}
\begin{tabular}{|c|l|c|c|c|c|c|}
\hline
  \multicolumn{1}{|c|}{\begin{tabular}[c]{@{}c@{}}Comp\\ Type\end{tabular}} &
  \multicolumn{1}{c|}{Env} &
  \multicolumn{1}{c|}{\begin{tabular}[c]{@{}c@{}}FPS\\ (base)\end{tabular}} &
  \multicolumn{1}{c|}{\begin{tabular}[c]{@{}c@{}}FPS\\ (ours)\end{tabular}} &
  \multicolumn{1}{c|}{\begin{tabular}[c]{@{}c@{}}GB\\ (base)\end{tabular}} &
  \multicolumn{1}{c|}{\begin{tabular}[c]{@{}c@{}}GB\\ (ours)\end{tabular}} &
  \multicolumn{1}{c|}{\begin{tabular}[c]{@{}c@{}}Total\\ Comp\end{tabular}} \\ \hline
        & Walker & 119 & 142 & 63.5 & 7.1 & 8.9$\times$ \\ 
Half & Quadruped & 184 & 243 & 63.6 & 7.1 & 9.0$\times$ \\ \hline
       & \textbf{AVG}  & \textbf{152} & \textbf{193} & \textbf{63.6} & \textbf{7.1} & \textbf{9.0$\times$} \\ \hline
        & Walker & 119 & 73 & 63.5 & 3.8 & 16.7$\times$ \\ 
Full & Quadruped & 184 & 135 & 63.6 & 4.1 & 15.5$\times$ \\ \hline
       & \textbf{AVG}  & \textbf{152} & \textbf{104} & \textbf{63.6} & \textbf{4.0} & \textbf{15.9$\times$} \\ \hline
\end{tabular}
\caption{DrQ-v2 training speed (FPS), replay buffer size (GB), and total memory reduction due to compression (Total Comp) of both speed-optimized (Half) and memory-optimized (Full) compression for the DMC~\cite{tassa2018deepmind} tasks of $\texttt{Walker Walk}$ and $\texttt{Quadruped Walk}$.}
\vspace{-10pt}
\label{table:dmc_stats}
\end{table}

\subsection{Robotic Control Tasks}

In this section, we evaluate our approach on the $\texttt{Walker Walk}$ and $\texttt{Quadruped Walk}$ tasks from the DMC robotic control suite~\cite{tassa2018deepmind}.\footnote{We make two minor modifications to the default environments: switching from RGB to grayscale image observations and removing the default checkerboard floor pattern.} We define two separate levels of image-based compression: Half and Full. Half, or speed-optimized compression uses a fully vectorized implementation of observation indexing (Section~\ref{sec:obs_ind}) but does not leverage observation compression (Section~\ref{sec:obs_comp}), while Full uses the entire compression stack detailed in Section~\ref{sec:method}, but as a result, does not support vectorization.\footnote{We note that this can be added through future work and as such speed results from Full compression represent a conservative underestimate of future fully-optimized performance.} In addition, both implementations store all experience values in our replay buffer directly in memory, unlike DrQ-v2, which by default needs to store such values to disk.

Table~\ref{table:dmc_stats} compares the training speed, replay buffer size, and total compression factor (Equation~\ref{eq:comp_factor}) achieved with Half and Full compression using DrQ-v2 across the $\texttt{Walker Walk}$ and $\texttt{Quadruped Walk}$ DMC tasks.\footnote{We set $N$ to be the maximum number of nonzero pixels stored at any time during training, across all trials, to provide a conservative bound.} We find that on average, Half and Full compression reduce the replay buffer's memory footprint by 9$\times$ and 15.9$\times$ respectively, with Full achieving a 16.7$\times$ compression factor for $\texttt{Walker Walk}$.

In terms of training speed, our Half compression approach, which is fully vectorized, improves latency over the baseline DrQ-v2 implementation by as much as 32\% while still achieving the aforementioned 9$\times$ compression factor. As noted earlier, our Full compression implementation does not yet leverage vectorization and so results in an average slow-down of 32\% as compared to the baseline DrQ-v2 implementation, which, despite storing its replay buffer on disk, leverages multi-threaded data loaders and batch pre-fetching in order to be as low latency as possible.

Figure~\ref{fig:dmc_curves} plots the learning curves of DrQ-v2 across the $\texttt{Walker Walk}$ and $\texttt{Quadruped Walk}$ DMC environments using no compression (black), Half compression (light blue), and Full compression (dark blue). We average our results across five random seeds and display the standard deviation with the accompanying shaded region. As expected, given that our compression is lossless, we find convergence is not affected by our approach, as all final rewards both with and without differential encoder-based compression are well within a standard deviation of each other.

\begin{figure}[!t]
     \centering
     \vspace{5pt}
     \includegraphics[width=0.48\textwidth]{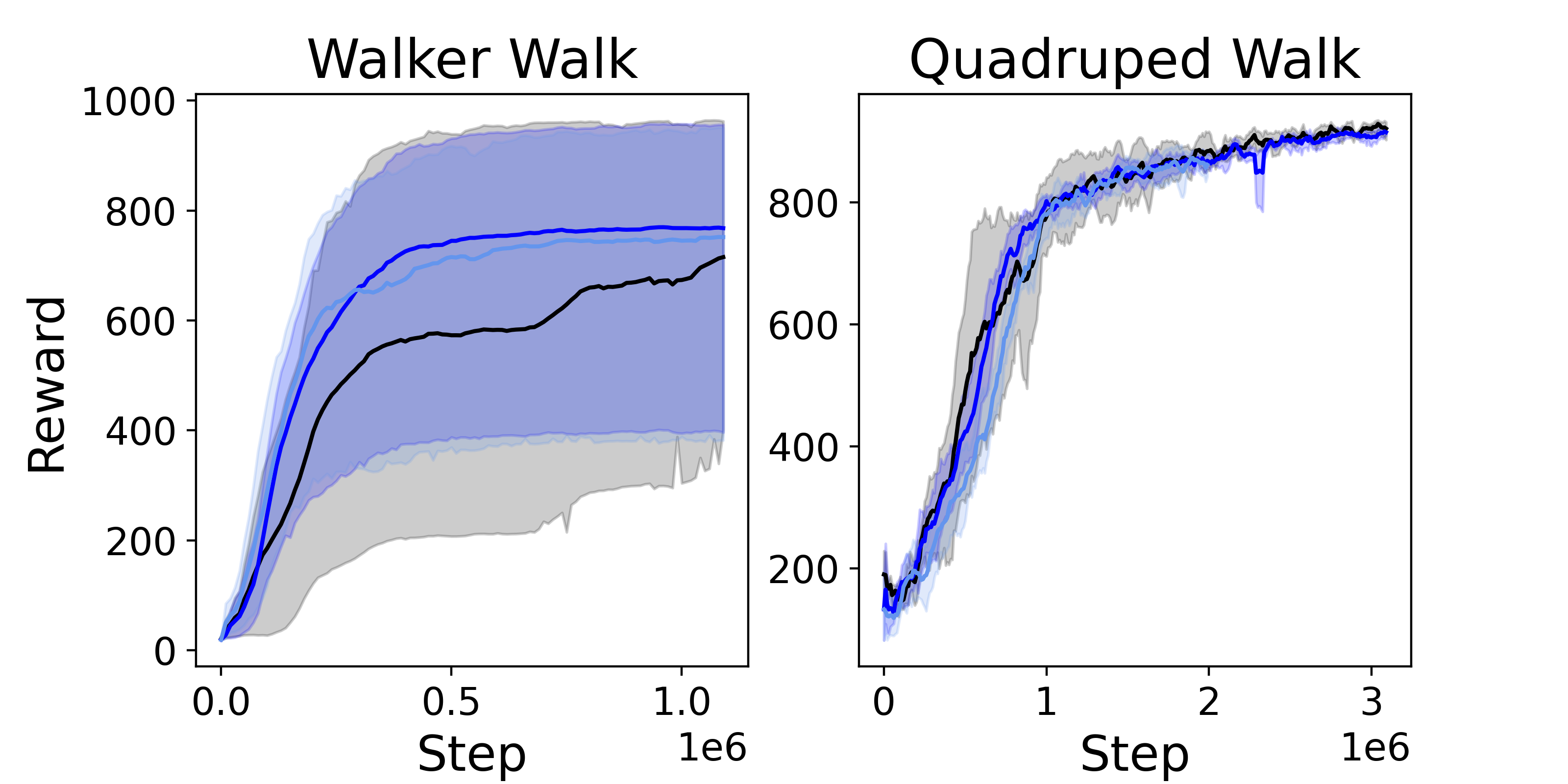}
     \caption{Learning curves of DrQ-v2 with Half (light blue), Full (dark blue), and no (black) compression for the DMC~\cite{tassa2018deepmind} tasks of $\texttt{Walker Walk}$ and $\texttt{Quadruped Walk}$.}
     \label{fig:dmc_curves}
     \vspace{-5pt}
\end{figure}

\subsection{Atari Environments} \label{sec:results:atari}

Table~\ref{table:atari_stats} compares the training speed and replay buffer size with and without differential encoding-based image compression across 10 Atari 2600 benchmark environments~\cite{Bellemare_2013ALE}, as well as the compression factor, for both PPO and QR-DQN using the Full compression approach.\footnote{We do not run the Atari experiments with Half compression, as due to their low memory cost relative to the DMC tasks, the Atari baselines are already fully vectorized in RAM.} We find that on average, Full compression is able to reduce the replay buffer's memory footprint by 8.9$\times$ and 9.9$\times$ for PPO and QR-DQN respectively. While the end compression factor does vary by environment--as the number $N$ of nonzero pixel values stored in the observation compressor (Equation~\ref{eq:obs_comp}) differs depending on the level of inter-frame dissimilarity within the environment--we find that 5/10 PPO and 6/10 QR-DQN environments achieve over a 10$\times$ compression factor, with our best-case environments achieving as much as a 14.2$\times$ reduction. As previously noted, Full compression is not fully vectorized and can be optimized through future work, as such we find an average 5\% and 29\% slow-down for PPO and QR-DQN respectively as compared to the default, vectorized, $\texttt{stable-baselines3}$ implementation.

Figure~\ref{fig:atari_curves} plots the learning curves of PPO and QR-DQN across the ten Atari environments for the first 10M steps of training. These figures report the average of five random seeds and also display the standard deviation with the accompanying shaded regions. As in the DMC environments, we find that convergence is not affected, with all final rewards being well within a standard deviation of one another. 

\begin{table}[!h]
\begin{tabular}{|c|l|c|c|c|c|c|}
\hline
\multicolumn{1}{|c|}{Alg} &
  \multicolumn{1}{c|}{Env} &
  \multicolumn{1}{c|}{\begin{tabular}[c]{@{}c@{}}FPS\\ (base)\end{tabular}} &
  \multicolumn{1}{c|}{\begin{tabular}[c]{@{}c@{}}FPS\\ (ours)\end{tabular}} &
  \multicolumn{1}{c|}{\begin{tabular}[c]{@{}c@{}}MB\\ (base)\end{tabular}} &
  \multicolumn{1}{c|}{\begin{tabular}[c]{@{}c@{}}MB\\ (ours)\end{tabular}} &
  \multicolumn{1}{c|}{\begin{tabular}[c]{@{}c@{}}Total\\ Comp\end{tabular}} \\ \hline
       & Asteroids     & 1605          & 1535          & 27.6          & 2.6          & 10.8$\times$         \\ 
       & BeamRider     & 1339          & 1267          & 27.6          & 7.6          & 3.6$\times$          \\ 
       & Breakout      & 1437          & 1383          & 27.6          & 2.0          & 14.2$\times$         \\ 
       & Enduro        & 1117          & 1062          & 27.6          & 3.5          & 7.9$\times$          \\ 
       & MsPacman      & 1432          & 1389          & 27.6          & 2.3          & 12.1$\times$         \\ 
PPO    & Pong          & 1582          & 1512          & 27.6          & 2.0          & 13.9$\times$         \\ 
       & Qbert         & 1507          & 1452          & 27.6          & 3.0          & 9.1$\times$          \\ 
       & RoadRunner    & 1347          & 1267          & 27.6          & 3.0          & 9.0$\times$          \\ 
       & Seaquest      & 1466          & 1371          & 27.6          & 2.9          & 9.6$\times$          \\ 
       & SpaceInvaders & 1468          & 1395          & 27.6          & 2.4          & 11.3$\times$         \\ \hline
       & \textbf{AVG}  & \textbf{1430} & \textbf{1363} & \textbf{27.6} & \textbf{3.1} & \textbf{8.9$\times$} \\ \hline
       & Asteroids     & 897           & 616           & 2694          & 265          & 10.2$\times$         \\ 
       & BeamRider     & 792           & 547           & 2694          & 552          & 4.9$\times$          \\ 
       & Breakout      & 833           & 587           & 2694          & 190          & 14.2$\times$         \\ 
       & Enduro        & 722           & 536           & 2694          & 306          & 8.8$\times$          \\ 
       & MsPacman      & 839           & 597           & 2694          & 219          & 12.3$\times$         \\ 
DQN & Pong          & 881           & 627           & 2694          & 192          & 14.0$\times$         \\ 
       & Qbert         & 849           & 608           & 2694          & 213          & 12.7$\times$         \\ 
       & RoadRunner    & 808           & 576           & 2694          & 290          & 9.3$\times$          \\ 
       & Seaquest      & 732           & 484           & 2694          & 278          & 9.7$\times$          \\ 
       & SpaceInvaders & 861           & 607           & 2694          & 227          & 11.9$\times$         \\ \hline
       & \textbf{AVG}  & \textbf{821}  & \textbf{579}  & \textbf{2694} & \textbf{273} & \textbf{9.9$\times$} \\ \hline
\end{tabular}
\caption{Training speed in frames/steps per second (FPS), replay buffer size (MB), and total memory reduction due to compression (Total Comp) of both PPO and QR-DQN (DQN) across ten Atari 2600~\cite{Bellemare_2013ALE} environments. }
\label{table:atari_stats}
\vspace{-10pt}
\end{table}
\fi

\section{CONCLUSION AND FUTURE WORK}
\ifevalnew
    \begin{figure*}[!t]
         \centering
         \vspace{5pt}
         \includegraphics[width=0.99\textwidth]{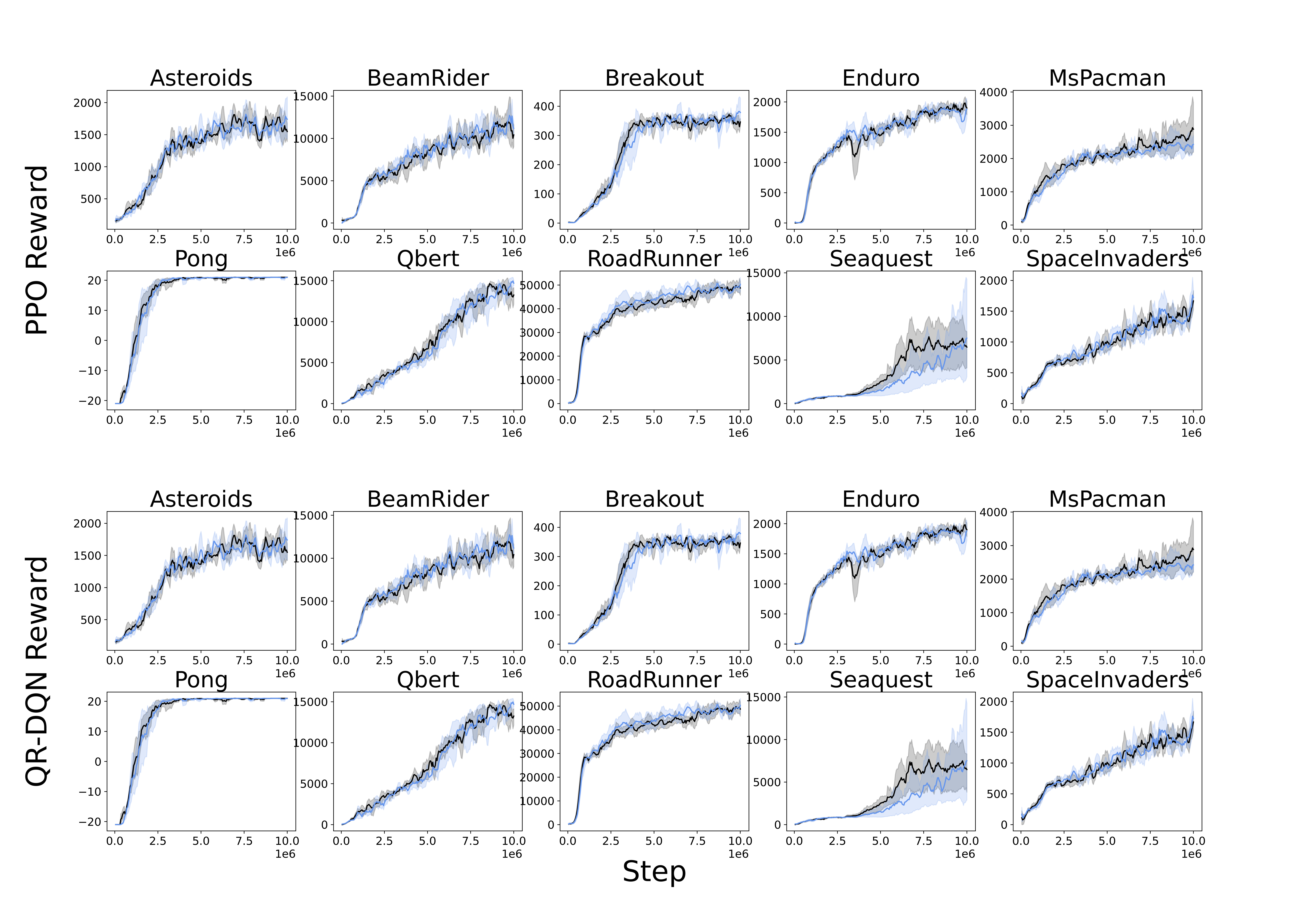}
         \vspace{-10pt}
         \caption{Learning curves of PPO (top) and QR-DQN (bottom) with (blue) and without (black) pixel-based compression across ten Atari 2600~\cite{Bellemare_2013ALE} environments. Rewards are averaged across five trials and standard deviations are shaded.}
         \label{fig:atari_curves}
    \end{figure*}
\fi

This paper presents a novel, differential encoding-based method for observation compression, reducing the overall memory requirements of perceptive DRL without impacting training performance.

We evaluate the compression factor, training speed, and learning performance across ten Atari and two DMC tasks using three state-of-the-art on- and off-policy perceptive DRL algorithms. 
We find that differential image encoding reduces the memory footprint by as much as 14.2$\times$ and 16.7$\times$ for the Atari and DMC tasks respectively. These savings also enable large-scale DRL that previously required paging between flash and RAM to be run entirely in RAM, improving latency for the DMC tasks by as much as 27\%.

Admittedly, not all learning-based approaches will benefit equally from observation compression. For instance, newer model-based techniques may trade off large replay buffers for more expressive world models~\cite{hafner2023mastering}. However, if we are to realize lifelong, practical learning on the edge, curbing memory usage, wherever it may be, is essential.

In future work, we hope to further optimize our method through enhanced vectorization and parallelization in order to speed up training. Finally, we hope to deploy our compression technique onto physical robot hardware and test it in the context of both real-world edge RL and tiny robot learning~\cite{Neuman22TinyRobotLearning}. We hope that this effort will aid in reducing cost and compute barriers for state-of-the-art RL across all robot platforms.

\bibliographystyle{inc/IEEEtran}
\bibliography{inc/main}

\begin{thebibliography}{10}
\providecommand{\url}[1]{#1}
\csname url@rmstyle\endcsname
\providecommand{\newblock}{\relax}
\providecommand{\bibinfo}[2]{#2}
\providecommand\BIBentrySTDinterwordspacing{\spaceskip=0pt\relax}
\providecommand\BIBentryALTinterwordstretchfactor{4}
\providecommand\BIBentryALTinterwordspacing{\spaceskip=\fontdimen2\font plus
\BIBentryALTinterwordstretchfactor\fontdimen3\font minus
  \fontdimen4\font\relax}
\providecommand\BIBforeignlanguage[2]{{%
\expandafter\ifx\csname l@#1\endcsname\relax
\typeout{** WARNING: IEEEtran.bst: No hyphenation pattern has been}%
\typeout{** loaded for the language `#1'. Using the pattern for}%
\typeout{** the default language instead.}%
\else
\language=\csname l@#1\endcsname
\fi
#2}}

\bibitem{Andrychowicz18}
\BIBentryALTinterwordspacing
{OpenAI}, M.~Andrychowicz, B.~Baker, M.~Chociej, R.~Jozefowicz, B.~McGrew,
  J.~Pachocki, A.~Petron, M.~Plappert, G.~Powell, A.~Ray, J.~Schneider,
  S.~Sidor, J.~Tobin, P.~Welinder, L.~Weng, and W.~Zaremba, ``Learning
  dexterous in-hand manipulation,'' 2018. [Online]. Available:
  \url{https://arxiv.org/abs/1808.00177}
\BIBentrySTDinterwordspacing

\bibitem{Miki22}
\BIBentryALTinterwordspacing
T.~Miki, J.~Lee, J.~Hwangbo, L.~Wellhausen, V.~Koltun, and M.~Hutter,
  ``Learning robust perceptive locomotion for quadrupedal robots in the wild,''
  \emph{Science Robotics}, vol.~7, no.~62, jan 2022. [Online]. Available:
  \url{https://doi.org/10.1126%2Fscirobotics.abk2822}
\BIBentrySTDinterwordspacing

\bibitem{Song21}
\BIBentryALTinterwordspacing
Y.~Song, M.~Steinweg, E.~Kaufmann, and D.~Scaramuzza, ``Autonomous drone racing
  with deep reinforcement learning,'' 2021. [Online]. Available:
  \url{https://arxiv.org/abs/2103.08624}
\BIBentrySTDinterwordspacing

\bibitem{silver2017mastering}
D.~Silver, T.~Hubert, J.~Schrittwieser, I.~Antonoglou, M.~Lai, A.~Guez,
  M.~Lanctot, L.~Sifre, D.~Kumaran, T.~Graepel, T.~Lillicrap, K.~Simonyan, and
  D.~Hassabis, ``Mastering chess and shogi by self-play with a general
  reinforcement learning algorithm,'' 2017.

\bibitem{openai2019dota}
OpenAI, :, C.~Berner, G.~Brockman, B.~Chan, V.~Cheung, P.~Dębiak, C.~Dennison,
  D.~Farhi, Q.~Fischer, S.~Hashme, C.~Hesse, R.~Józefowicz, S.~Gray,
  C.~Olsson, J.~Pachocki, M.~Petrov, H.~P. d.~O.~Pinto, J.~Raiman, T.~Salimans,
  J.~Schlatter, J.~Schneider, S.~Sidor, I.~Sutskever, J.~Tang, F.~Wolski, and
  S.~Zhang, ``Dota 2 with large scale deep reinforcement learning,'' 2019.

\bibitem{hafner2023mastering}
D.~Hafner, J.~Pasukonis, J.~Ba, and T.~Lillicrap, ``Mastering diverse domains
  through world models,'' 2023.

\bibitem{grossman23JustRound}
L.~Grossman and B.~Plancher, ``Just round: Quantized observation spaces enable
  memory efficient learning of dynamic locomotion,'' in \emph{IEEE
  International Conference on Robotics and Automation (ICRA)}, London, UK, May.
  2023.

\bibitem{yarats2021DrQ-v2}
D.~Yarats, R.~Fergus, A.~Lazaric, and L.~Pinto, ``Mastering visual continuous
  control: Improved data-augmented reinforcement learning,'' 2021.

\bibitem{tassa2018deepmind}
Y.~Tassa, Y.~Doron, A.~Muldal, T.~Erez, Y.~Li, D.~de~Las~Casas, D.~Budden,
  A.~Abdolmaleki, J.~Merel, A.~Lefrancq, T.~Lillicrap, and M.~Riedmiller,
  ``Deepmind control suite,'' 2018.

\bibitem{Gankidi17}
P.~R. Gankidi and J.~Thangavelautham, ``Fpga architecture for deep learning and
  its application to planetary robotics,'' in \emph{2017 IEEE Aerospace
  Conference}, 2017, pp. 1--9.

\bibitem{Thrun95}
\BIBentryALTinterwordspacing
S.~Thrun and T.~M. Mitchell, ``Lifelong robot learning,'' \emph{Robotics and
  Autonomous Systems}, vol.~15, no.~1, pp. 25--46, 1995, the Biology and
  Technology of Intelligent Autonomous Agents. [Online]. Available:
  \url{https://www.sciencedirect.com/science/article/pii/092188909500004Y}
\BIBentrySTDinterwordspacing

\bibitem{Bellemare_2013ALE}
\BIBentryALTinterwordspacing
M.~G. Bellemare, Y.~Naddaf, J.~Veness, and M.~Bowling, ``The arcade learning
  environment: An evaluation platform for general agents,'' \emph{Journal of
  Artificial Intelligence Research}, vol.~47, pp. 253--279, jun 2013. [Online].
  Available: \url{https://doi.org/10.1613%2Fjair.3912}
\BIBentrySTDinterwordspacing

\bibitem{mnih2013playing}
V.~Mnih, K.~Kavukcuoglu, D.~Silver, A.~Graves, I.~Antonoglou, D.~Wierstra, and
  M.~Riedmiller, ``Playing atari with deep reinforcement learning,'' 2013.

\bibitem{hessel2017rainbow}
M.~Hessel, J.~Modayil, H.~van Hasselt, T.~Schaul, G.~Ostrovski, W.~Dabney,
  D.~Horgan, B.~Piot, M.~Azar, and D.~Silver, ``Rainbow: Combining improvements
  in deep reinforcement learning,'' 2017.

\bibitem{dabney2017distributional}
W.~Dabney, M.~Rowland, M.~G. Bellemare, and R.~Munos, ``Distributional
  reinforcement learning with quantile regression,'' 2017.

\bibitem{kaiser2020modelbasedSimPLe}
L.~Kaiser, M.~Babaeizadeh, P.~Milos, B.~Osinski, R.~H. Campbell, K.~Czechowski,
  D.~Erhan, C.~Finn, P.~Kozakowski, S.~Levine, A.~Mohiuddin, R.~Sepassi,
  G.~Tucker, and H.~Michalewski, ``Model-based reinforcement learning for
  atari,'' 2020.

\bibitem{hafner2019learningPlaNet}
D.~Hafner, T.~Lillicrap, I.~Fischer, R.~Villegas, D.~Ha, H.~Lee, and
  J.~Davidson, ``Learning latent dynamics for planning from pixels,'' 2019.

\bibitem{srinivas2020curl}
A.~Srinivas, M.~Laskin, and P.~Abbeel, ``Curl: Contrastive unsupervised
  representations for reinforcement learning,'' 2020.

\bibitem{kostrikov2021DrQ}
I.~Kostrikov, D.~Yarats, and R.~Fergus, ``Image augmentation is all you need:
  Regularizing deep reinforcement learning from pixels,'' 2021.

\bibitem{laskin2020reinforcementaugment}
M.~Laskin, K.~Lee, A.~Stooke, L.~Pinto, P.~Abbeel, and A.~Srinivas,
  ``Reinforcement learning with augmented data,'' 2020.

\bibitem{yarats2021reinforcementrepresentations}
D.~Yarats, R.~Fergus, A.~Lazaric, and L.~Pinto, ``Reinforcement learning with
  prototypical representations,'' 2021.

\bibitem{laskin2020reinforcementgrayscale}
M.~Laskin, K.~Lee, A.~Stooke, L.~Pinto, P.~Abbeel, and A.~Srinivas,
  ``Reinforcement learning with augmented data,'' 2020.

\bibitem{Hafner18}
\BIBentryALTinterwordspacing
D.~Hafner, T.~Lillicrap, I.~Fischer, R.~Villegas, D.~Ha, H.~Lee, and
  J.~Davidson, ``Learning latent dynamics for planning from pixels,'' 2018.
  [Online]. Available: \url{https://arxiv.org/abs/1811.04551}
\BIBentrySTDinterwordspacing

\bibitem{hafner2022mastering}
D.~Hafner, T.~Lillicrap, M.~Norouzi, and J.~Ba, ``Mastering atari with discrete
  world models,'' 2022.

\bibitem{wu2022daydreamer}
P.~Wu, A.~Escontrela, D.~Hafner, K.~Goldberg, and P.~Abbeel, ``Daydreamer:
  World models for physical robot learning,'' 2022.

\bibitem{tomar2021learning}
M.~Tomar, U.~A. Mishra, A.~Zhang, and M.~E. Taylor, ``Learning representations
  for pixel-based control: What matters and why?'' 2021.

\bibitem{pmlr-v164-yu22a}
\BIBentryALTinterwordspacing
W.~Yu, D.~Jain, A.~Escontrela, A.~Iscen, P.~Xu, E.~Coumans, S.~Ha, J.~Tan, and
  T.~Zhang, ``Visual-locomotion: Learning to walk on complex terrains with
  vision,'' in \emph{Proceedings of the 5th Conference on Robot Learning}, ser.
  Proceedings of Machine Learning Research, A.~Faust, D.~Hsu, and G.~Neumann,
  Eds., vol. 164.\hskip 1em plus 0.5em minus 0.4em\relax PMLR, 08--11 Nov 2022,
  pp. 1291--1302. [Online]. Available:
  \url{https://proceedings.mlr.press/v164/yu22a.html}
\BIBentrySTDinterwordspacing

\bibitem{wood2022task}
D.~Wood, ``Task oriented video coding: A survey,'' 2022.

\bibitem{MPEG-4}
\BIBentryALTinterwordspacing
T.~Ebrahimi and C.~Horne, ``Mpeg-4 natural video coding – an overview,''
  \emph{Signal Processing: Image Communication}, vol.~15, no.~4, pp. 365--385,
  2000. [Online]. Available:
  \url{https://www.sciencedirect.com/science/article/pii/S0923596599000545}
\BIBentrySTDinterwordspacing

\bibitem{chen2006introductionH264}
J.-W. Chen, C.-Y. Kao, and Y.-L. Lin, ``Introduction to h. 264 advanced video
  coding,'' in \emph{Proceedings of the 2006 Asia and South Pacific Design
  Automation Conference}, 2006, pp. 736--741.

\bibitem{yang2021videoVCM}
W.~Yang, H.~Huang, Y.~Hu, L.-Y. Duan, and J.~Liu, ``Video coding for machine:
  Compact visual representation compression for intelligent collaborative
  analytics,'' 2021.

\bibitem{huang2021visualVCSclassification}
Z.~Huang, C.~Jia, S.~Wang, and S.~Ma, ``Visual analysis motivated
  rate-distortion model for image coding,'' 2021.

\bibitem{Yang_2020VCSdetection}
\BIBentryALTinterwordspacing
Z.~Yang, Y.~Wang, C.~Xu, P.~Du, C.~Xu, C.~Xu, and Q.~Tian, ``Discernible image
  compression,'' in \emph{Proceedings of the 28th {ACM} International
  Conference on Multimedia}.\hskip 1em plus 0.5em minus 0.4em\relax {ACM}, oct
  2020. [Online]. Available: \url{https://doi.org/10.1145%2F3394171.3413968}
\BIBentrySTDinterwordspacing

\bibitem{Le_2021VCSInstance}
\BIBentryALTinterwordspacing
N.~Le, H.~Zhang, F.~Cricri, R.~Ghaznavi-Youvalari, and E.~Rahtu, ``Image coding
  for machines: an end-to-end learned approach,'' in \emph{{ICASSP} 2021 - 2021
  {IEEE} International Conference on Acoustics, Speech and Signal Processing
  ({ICASSP})}.\hskip 1em plus 0.5em minus 0.4em\relax {IEEE}, jun 2021.
  [Online]. Available: \url{https://doi.org/10.1109%2Ficassp39728.2021.9414465}
\BIBentrySTDinterwordspacing

\bibitem{Schulman17}
J.~Schulman, F.~Wolski, P.~Dhariwal, A.~Radford, and O.~Klimov, ``Proximal
  {{Policy Optimization Algorithms}},'' \emph{arXiv:1707.06347 [cs]}, July
  2017.

\bibitem{Sutton00}
R.~S. Sutton, D.~M.~A. {llester}, S.~Singh, and Y.~Mansour, ``Policy {{Gradient
  Methods}} for {{Reinforcement Learning}} with {{Function Approximation}},''
  2000, pp. 1057--1063.

\bibitem{huber64}
\BIBentryALTinterwordspacing
P.~J. Huber, ``{Robust Estimation of a Location Parameter},'' \emph{The Annals
  of Mathematical Statistics}, vol.~35, no.~1, pp. 73 -- 101, 1964. [Online].
  Available: \url{https://doi.org/10.1214/aoms/1177703732}
\BIBentrySTDinterwordspacing

\bibitem{Lillicrap15}
\BIBentryALTinterwordspacing
T.~P. Lillicrap, J.~J. Hunt, A.~Pritzel, N.~Heess, T.~Erez, Y.~Tassa,
  D.~Silver, and D.~Wierstra, ``Continuous control with deep reinforcement
  learning,'' 2015. [Online]. Available: \url{https://arxiv.org/abs/1509.02971}
\BIBentrySTDinterwordspacing

\bibitem{IntroDataCompCh11DE}
\BIBentryALTinterwordspacing
K.~Sayood, ``Chapter 11 - differential encoding,'' in \emph{Introduction to
  Data Compression (Fifth Edition)}, fifth edition~ed., ser. The Morgan
  Kaufmann Series in Multimedia Information and Systems, K.~Sayood, Ed.\hskip
  1em plus 0.5em minus 0.4em\relax Morgan Kaufmann, 2018, pp. 351--378.
  [Online]. Available:
  \url{https://www.sciencedirect.com/science/article/pii/B9780128094747000112}
\BIBentrySTDinterwordspacing

\bibitem{ruparelia2010historyVCS}
N.~B. Ruparelia, ``The history of version control,'' \emph{ACM SIGSOFT Software
  Engineering Notes}, vol.~35, no.~1, pp. 5--9, 2010.

\bibitem{Lin92ExperienceReplay}
\BIBentryALTinterwordspacing
L.-J. Lin, ``Self-improving reactive agents based on reinforcement learning,
  planning and teaching,'' \emph{Machine Learning}, vol.~8, no.~3, pp.
  293--321, 1992. [Online]. Available: \url{https://doi.org/10.1007/BF00992699}
\BIBentrySTDinterwordspacing

\bibitem{schaul2016prioritized}
T.~Schaul, J.~Quan, I.~Antonoglou, and D.~Silver, ``Prioritized experience
  replay,'' 2016.

\bibitem{stable-baselines3}
\BIBentryALTinterwordspacing
A.~Raffin, A.~Hill, A.~Gleave, A.~Kanervisto, M.~Ernestus, and N.~Dormann,
  ``Stable-baselines3: Reliable reinforcement learning implementations,''
  \emph{Journal of Machine Learning Research}, vol.~22, no. 268, pp. 1--8,
  2021. [Online]. Available: \url{http://jmlr.org/papers/v22/20-1364.html}
\BIBentrySTDinterwordspacing

\bibitem{Neuman22TinyRobotLearning}
S.~M. Neuman, B.~Plancher, B.~P. Duisterhof, S.~Krishnan, C.~Banbury,
  M.~Mazumder, S.~Prakash, J.~Jabbour, A.~Faust, G.~C. de~Croon, and
  V.~Janapa~Reddi, ``Tiny robot learning: Challenges and directions for machine
  learning in resource-constrained robots,'' in \emph{2022 IEEE International
  Conference on Artificial Intelligence Circuits and Systems (AICAS)}, Incheon,
  Korea, June 2022.

\end{thebibliography}

\end{document}